\documentclass[journal]{IEEEtran}
\usepackage{amsmath}
\usepackage{mathtools}
\usepackage{cases}
\usepackage{esvect}
\usepackage[algo2e]{algorithm2e}
\usepackage{graphicx}
\usepackage{caption}
\usepackage{subfig}
\usepackage{lipsum}
\usepackage{multicol}
\usepackage{enumerate}
\usepackage{algorithm}
\usepackage{hyperref}
\usepackage{algorithmicx}
\usepackage{algorithm,algpseudocode}
\usepackage{float}
\usepackage{booktabs}
\usepackage[export]{adjustbox}
\usepackage{soul,color}
\usepackage{multirow}
\usepackage{textcomp}
\usepackage{comment}
\begin{document}
\title{DRBM-ClustNet: A Deep Restricted Boltzmann-Kohonen Architecture for Data Clustering}
\markboth{Preprint version}%
{Shell \MakeLowercase{\textit{et al.}}: Bare Demo of IEEEtran.cls for IEEE Journals}

\author{J. Senthilnath, Senior Member, IEEE, Nagaraj G, Sumanth Simha C, Sushant Kulkarni, Meenakumari Thapa, Indiramma M and J\'{o}n Atli Benediktsson, Fellow, IEEE

\thanks{J. Senthilnath is with the Institute for Infocomm Research, Agency for Science, Technology and Research (A*STAR), 138632 Singapore (e-mail: J\_Senthilnath@i2r.a$-$star.edu.sg).}
\thanks{Nagaraj G is with the B.M.S. College of Engineering, Bengaluru 560019, India (e-mail: nagaraj170297@gmail.com).}
\thanks{Sumanth Simha C is with the B.M.S. College of Engineering, Bengaluru 560019, India (e-mail: ssimha152@gmail.com).}
\thanks{Sushant Kulkarni is with the Indian Institute of Technology, Chennai 600036, India (e-mail: ksushantk@gmail.com).}
\thanks{Meenakumari Thapa is with the Birla Institute of Technology and Science, Pilani 333031, India (e-mail: meenathapa31@gmail.com).}
\thanks{Indiramma M is  with the B.M.S. College of Engineering, Bengaluru 560019, India (e-mail: indira.cse@bmsce.ac.in).}
\thanks{J\'{o}n Atli Benediktsson is with the Faculty of Electrical  and Computer Engineering, University of Iceland, 107 Reykjavik, Iceland (e-mail: benedikt@hi.is).}
}
\maketitle
\begin{abstract}
A Bayesian Deep Restricted Boltzmann-Kohonen architecture for data clustering termed as DRBM-ClustNet is proposed. This core-clustering engine consists of a Deep Restricted Boltzmann Machine (DRBM) for processing unlabeled data by creating new features that are uncorrelated and have large variance with each other. Next, the number of clusters are predicted using the Bayesian Information Criterion (BIC), followed by a Kohonen Network-based clustering layer. The processing of unlabeled data is done in three stages for efficient clustering of the non-linearly separable datasets. In the first stage, DRBM performs non-linear feature extraction by capturing the highly complex data representation by projecting the feature vectors of $d$ dimensions into $n$ dimensions. Most clustering algorithms require the number of clusters to be decided a priori, hence here to automate the number of clusters in the second stage we use BIC. In the third stage, the number of clusters derived from BIC forms the input for the Kohonen network, which performs clustering of the feature-extracted data obtained from the DRBM. This method overcomes the general disadvantages of clustering algorithms like the prior specification of the number of clusters, convergence to local optima and poor clustering accuracy on non-linear datasets. In this research we use two synthetic datasets, fifteen benchmark datasets from the UCI Machine Learning repository, and four image datasets to analyze the DRBM-ClustNet. The proposed framework is evaluated based on clustering accuracy and ranked against other state-of-the-art clustering methods. The obtained results demonstrate that the DRBM-ClustNet outperforms state-of-the-art clustering algorithms.
\end{abstract}

\graphicspath{{Figures/}}
\IEEEpeerreviewmaketitle

\section{Introduction}
\IEEEPARstart Clustering is the task of the division of data into groups of similar objects \cite{n1}. The main objective of a clustering algorithm is to explore the inherent structure present in the samples and group the related samples \cite{n2}. Clustering is very useful in organizing data and understanding the hidden structure of the data. Clustering finds its plethora of applications in text analytics \cite{n3} \cite{n3a}, anomaly detection \cite{n4}, hand-written digit recognition \cite{n5}, social network analysis \cite{n6}, customer grouping based on preferences in marketing sites \cite{n7}, dynamic clustering \cite{n8}, face recognition \cite{n9}, model order detection \cite{n9a} and heterogeneous system placement problems \cite{n10}.

There are various types of clustering algorithms such as partitional clustering, hierarchical clustering, model-based clustering, and density-based clustering \cite{n11}. Most popular partitional clustering algorithms are K-means and the Kohonen Network. In the case of model-based clustering, the expectation maximization (EM) Algorithm is most popular \cite{n12} \cite{n13} \cite{n14}. For K-means, the cluster centers are initialized randomly, and the samples are assigned to any one of the clusters using a similarity measure \cite{n12}. In every iteration, these cluster centers are updated until convergence is reached. In the Kohonen Network, the attributes that are assigned weights are updated iteratively using a competitive learning mechanism \cite{n13}. The EM algorithm groups the data based on the maximum likelihood approach using the Gaussian mixture model \cite{n14}. The main drawback of the aforementioned clustering algorithms is that their performance mainly depends on the dimensionality of the feature vector and the prior assignment of the number of clusters \cite{n15} \cite{n16}. By using the Bayesian Information Criterion (BIC), we can remove the problem of specifying the number of clusters a priori \cite{n17}.

A common misconception in clustering is that having a larger number of samples with cluster information leads to a better discriminative cluster since the features contain information about the target (class/cluster). However, there are many reasons like the presence of irrelevant, redundant or noisy features, which thwarts the clustering accuracy. Thus, it is essential to project features into a higher or lower order (and vice versa) by transforming non-linearly separable data into linearly separable data. In the literature, the Restricted Boltzmann Machine (RBM) has overcome such issues \cite{n18} \cite{n19} \cite{n20}. The RBM is a non-linear feature extraction technique that is superior in comparison to linear feature extraction techniques such as Principal Component Analysis \cite{n21}.

The first Boltzmann Machine (BM) model was developed in the 1980s and was inspired by statistical mechanics \cite{n22}. Smolensky in 1986 \cite{n23} implemented the Harmonium, which is a variant of the BM. The RBM is often described as the Monte Carlo version of Hopfield networks \cite{n24}.  Even though the idea of BM has existed since the 1980s, it was more widely used from 2006 when Hinton implemented a modified version of BM, termed as RBM, for a fast and efficient way of learning \cite{n25a}. RBM overcomes problems with the time required for the network to converge, which grows exponentially with the size of the samples \cite{n26}. In BM, it is difficult to obtain an unbiased sample from the posterior distribution given the data vector; the complexity is due to the connectivity within and between the layers. In RBM, the $restrictedness$ arises due to the absence of visible to visible and hidden to hidden interactions in the network. 

The learning algorithm used to train the RBM is based on a $k$-step contrastive divergence (CD) technique, which approximates the Markov Chain Monte Carlo (MCMC) technique by sampling all the units in a layer at once \cite{n27}. After the emergence of this technique, and with advances in the availability of computational power, many researchers realized that shallow neural networks were not sufficient. Thus, there was a need for networks to learn deeper representations of the data \cite{n28} which is essential for computer vision \cite{n29}, time-series forecasting and other applications \cite{n30}. Later, Hinton et al., \cite{n31} developed Deep Boltzmann Machines by stacking many layers of RBM and forming an undirected graphical model, which is unlike Deep Belief Networks, and uses directed graphical models \cite{n25a} \cite{n25b}. The Deep Restricted Boltzmann Machine (DRBM) has been used as a generative model to learn the non-linear distribution of a dataset \cite{n30}. The development of better computational hardware and high-performance processors has recently led to applications  of the approach with larger datasets \cite{n30a} \cite{n30b}. Sankaran et al., \cite{n32} proposed an RBM-based semi-supervised class sparsity signature by combining the unsupervised generative training with a supervised sparsity regularizer to learn better generative features. The RBM has also been applied for cancer detection \cite{n33}, time-series prediction \cite{new11}, detecting event-related potential \cite{new1}, and estimation of 3D trajectories from 2D trajectories \cite{n34}.

In this paper, we propose the DRBM-ClustNet framework, which performs feature extraction using DRBM. The output of the first stage is used to predict the number of clusters with BIC. The feature-extracted data obtained with DRBM and the predicted number of clusters are used as an input for the clustering algorithm using the Kohonen Network. The proposed framework's performance is compared with six state-of-the-art clustering methods, namely, K-means \cite{n35}, Self-Organizing Maps (SOM) \cite{n36}, Expectation Maximization (EM) \cite{n37}, Density-Based Spatial Clustering of Applications with Noise (DBSCAN) \cite{nw3}, Unsupervised Extreme Learning Machine (US-ELM) \cite{n38}, and Bayesian Extreme Learning Machines Kohonen Network (BELMKN) \cite{n39}. The proposed DRBM-ClustNet and the state-of-the-art clustering algorithms are applied on 2 synthetic datasets, 15 benchmark datasets from the UCI Machine Learning Repository \cite{n40}, and 4 image datasets. The performance evaluation is carried out using various statistical approaches. 

The remaining content of the paper is organized as follows: The architecture diagram of the DRBM-ClustNet framework with an abstract code is discussed in Section II. The clustering performance for 2 synthetic datasets with the DRBM-ClustNet framework and other available clustering algorithms is narrated in Section III. The results for various clustering methods when applied on 15 benchmark tabular datasets and 4 image datasets compared to the proposed approach are discussed in Section IV. Conclusions are drawn in Section V.

\section{DRBM-ClustNet Architecture}
\begin{figure*}[!htbp]
    \centering
        \captionsetup{justification=centering}
    \includegraphics[width=0.8\textwidth]{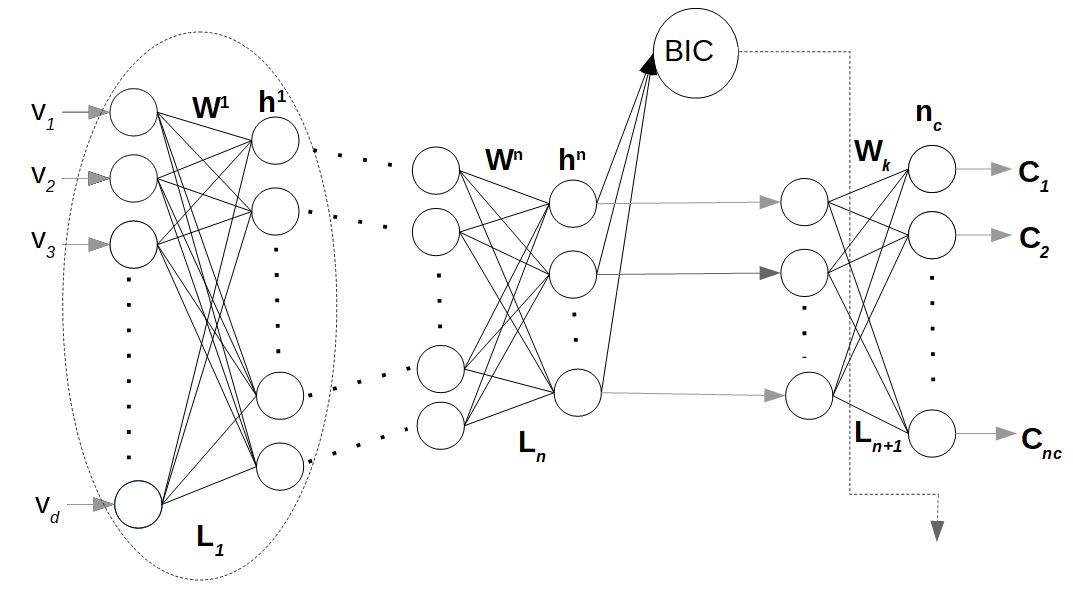}
    \caption{DRBM-ClustNet architecture}
    \label{ClusArch}
\end{figure*}

The proposed DRBM-ClustNet architecture for data clustering is discussed in this section. The DRBM is used for feature extraction, followed by BIC which is applied to predict the number of cluster centers for the feature extracted data. Finally, the above two levels are used as input for data clustering using the Kohonen Network. The detailed flow of the DRBM-ClustNet architecture is shown in Fig. \ref{ClusArch}.

\subsection{Feature extraction using DRBM}
The RBM is an energy-based undirected bipartite graphical model as shown in layer 1 ($\boldsymbol{L_1}$) of Fig. \ref{ClusArch}. For a given dataset, initially, we use the min-max normalization on a real-valued dataset for feature extraction, whereas the conventional RBM uses binary-valued data. The normalization of the dataset performed is the same as in \cite{nw1}, \cite{nw2}

\begin{equation}
    F(\textbf{x}_s) = \frac{0.9-0.1}{max(\textbf{x}_i) - min(\textbf{x}_i)}(\textbf{x}_s - min(\textbf{x}_i))+0.1,
\label{eq:eq1}
\end{equation}

\noindent
where $\textbf{x}_s$ is a dataset with $s=\{1,2,...,N\}$, $min(\textbf{x}_i)$ and $max(\textbf{x}_i)$ corresponds to the minimum and maximum value for a particular feature with $i=\{1,2,...,d\}$. 

The normalized dataset is passed to $\boldsymbol{L_1}$ of the DRBM network but $\boldsymbol{L_1}$ consists of a visible layer and a hidden layer. The visible units $v_i$, $i \in $ $1$ to $d$ are real-valued and the hidden units are given by $h_j$, $j \in$ 1 to $p$ . Let $W^1 \in 
{\rm I\!R}^{d\times p}$ denote the weights between the hidden layer and visible layer units in $\boldsymbol{L_1}$ and $w_{ij}$ indicate the weights between the $i^{th}$ visible unit and the $j^{th}$ hidden unit. As RBM is an energy-based model, the probability distribution of the visible layer and hidden layer units are stated in the form of an energy equation. The joint probability distribution of the visible layer and the hidden layer units $P(v,h)$ is given by

\begin{equation}
    P(v,h) = \frac{1}{Z}e^{-E(v,h)},
\label{eq:eq2}
\end{equation}

\noindent
where $E(v,h)$ is the energy of the joint configuration between visible layer and hidden layer units and is given by

\begin{equation}
    E(v,h) = -\sum_{i=1}^{d}\sum_{j=1}^{p}w_{ij}v_ih_j - \sum_{i=1}^{d}b_iv_i - \sum_{j=1}^{p}c_jh_j,
\label{eq:eq3}
\end{equation}

\noindent
where $b_i$ is the visible layer bias and $c_j$ is the hidden layer bias. $Z$ is the partition function defined as

\begin{equation}
    Z = \sum_{i=1}^{d}\sum_{j=1}^{p}e^{-E(v_i,h_j)}.
\label{eq:eq4}
\end{equation}

The partition function $(Z)$ is intractable. As a result, the joint probability distribution of the visible and the hidden units $P(v,h)$ which depends on $Z$ is also intractable. Therefore, we express the joint probability distribution $P(v,h)$ in (\ref{eq:eq2}) in terms of the conditional probability distribution $P(v|h)$ and $P(h|v)$. This is much easier to compute and is defined as

\begin{equation}
    P(v|h) = \prod_{i=1}^{d}P(v_i|h),
\label{eq:eq5}
\end{equation}

\begin{equation}
    P(h|v) = \prod_{j=1}^{p}P(h_j|v).
\label{eq:eq6}
\end{equation}

From (\ref{eq:eq5}) and (\ref{eq:eq6}) we can observe that the conditional probabilities are all independent. This indicates that the state of the hidden layer units is independent given the states of the visible layer units (without interconnecting the neurons in the same layer either in visible or hidden layer). Therefore, this type of Boltzmann Machine is defined as a Restricted Boltzmann Machine (RBM). The conditional probabilities from (\ref{eq:eq5}) and (\ref{eq:eq6}) are derived as a sigmoidal function given by

\begin{equation}
    P(h_j|v) = s(\sum_{i=1}^{d}v_iw_{ij} + c_j),
\label{eq:eq7}
\end{equation}

\begin{equation}
    P(v_i|h) = s(\sum_{j=1}^{p}h_jw_{ij} + b_i),
\label{eq:eq8}
\end{equation}

\noindent
where $s(.)$ denotes the sigmoidal function given by $s(x) = \frac{1}{1+e^{-x}}$.

RBM is a generative model from which we predict $P(h|v)$ by clamping the visible units and $P(v|h)$ by using the hidden activation obtained from $P(h|v)$. Hinton proposed the \emph{k}-step Contrastive Divergence algorithm (CD) \cite{n27} for training the RBM, which jointly performs the Gibbs Sampling for all variables in one layer instead of sampling the new values of all variables one-by-one. The one-step CD is divided into two phases namely, positive contrastive divergence and negative contrastive divergence followed by weight updating.

\begin{enumerate}
    \item \emph{Positive contrastive divergence}: The probability that a hidden state is evaluated by clamping all the visible units:
    \begin{equation}
        P(h|v) = s(v_{<0>}W+c) = h_{<0>}.
    \label{eq:eq9}
    \end{equation}
    Then, we calculate the interactions between the visible layer and hidden layer units as
    \begin{equation}
        a^+ = v_{<0>}h_{<0>} = (vh)_{<0>},
    \label{eq:eq10}
    \end{equation}
    where $a^+$  denotes the positive associations.
    
    \item \emph{Negative contrastive divergence}: The probability of a visible state given the hidden activations from (\ref{eq:eq9}) to get $v_{<1>}$. Then we use these visible units to find $h_{<1>}$:
    \begin{equation}
        P(v|h) = s(Wh_{<0>}+b) = v_{<1>},
    \label{eq:eq11}
    \end{equation}
    \begin{equation}
        P(h|v) = s(Wv_{<1>}+c) = h_{<1>}.
    \label{eq:eq12}
    \end{equation}
    Then, we calculate the negative associations between the visible layer and hidden layer units
    \begin{equation}
        a^- = v_{<1>}h_{<1>} = (vh)_{<1>}.
    \label{eq:eq13}
    \end{equation}
    
    \item \emph{Weight update step}: Update the weights by considering the difference between the positive associations from (\ref{eq:eq10}) and negative associations from (\ref{eq:eq13}) using adaptively decreasing learning rate $(\epsilon)$
    \begin{equation}
        \Delta w = \epsilon(a^+ - a^-) .
    \label{eq:eq14}
    \end{equation}
\end{enumerate}{}

The above one-step CD algorithm is applied and the states of visible and hidden units are sampled alternately.

In many cases, a single layer ($\boldsymbol{L_1}$) RBM will not be able to capture the complete non-linearity in the dataset. As a result, we need to stack multiple RBM layers ($\boldsymbol{L_1}$, $\boldsymbol{L_2}$,..., $\boldsymbol{L_n}$) to produce a DRBM. In DRBM, the activation of one hidden layer forms the training samples for the next hidden layer. This technique is efficient in learning the complex data representations. DRBM can efficiently learn a generative model of high dimensional input and large-scale dataset \cite{n18}.

Consider the DRBM network in the proposed architecture shown in Fig. \ref{ClusArch}, where feature extraction is carried out using the layers ($\boldsymbol{L_1}$, $\boldsymbol{L_2}$,...,$\boldsymbol{L_n}$). Let $h^l$ $(l\in$ 1 to $n$) denote the $l^{th}$ hidden layer, $W^l$ denote the weights between the layers. The energy equation for the DRBM is given by

\begin{equation}
    E(v,h^1,..,h^n;\theta) = -vW^1h^1-h^1W^2h^2-..-h^{n-1}W^nh^n,
\label{eq:eq15}
\end{equation}

\noindent
where $\theta = \{W^1,W^2,...,W^n\}$ are the model parameters which have to be trained and $n$ is the number of hidden layers in the DRBM network \cite{n41}.

In DRBM, the first layer RBM is trained using one-step contrastive divergence to obtain the reconstructions of the visible vectors. Then,  the next layer RBM is trained using the first layer sampled hidden layer activations $h^l$ obtained from $P(h^l|v;W^l)$ as the training data for the second layer RBM and obtain the reconstructions of the visible units. Continue this process recursively until layers $n-1$. Using all the trained weights $\theta = \{W^1,W^2,...,W^n\}$  and the input data features as the visible units, a feedforward pass is performed until the last hidden layer $n$. The final activations of the hidden layer $n$ are the feature extracted data ($\tilde{\textbf{x}}_{s}^{l}$).

\subsection{BIC for Cluster Prediction}
To automate the process of cluster prediction, DRBM-ClustNet uses the BIC. This is applied to the feature extracted data obtained from DRBM instead of the actual dataset as shown in Fig. \ref{ClusArch}. Most of the clustering algorithms require a parameter called the number of clusters to be inputted manually. As in real-world problems, datasets may be with or without labels, hence BIC is applied to statistically predict the number of clusters.

BIC uses multivariate Gaussian distribution and parameters, which are the mean and covariance matrices. These parameters are estimated using the EM algorithm. BIC for cluster prediction is defined as
\begin{equation}
    BIC = ln(N)\emph{k} - 2ln(\Tilde{L}),
\label{eq:eq16}
\end{equation}

\noindent
where $N$ is the total number of samples, $\Tilde{L}$ is the maximized value of the likelihood function and $k$ is the total number of free parameters to be estimated. The BIC is computed for $c=1,2,...,N$ and the model value with minimum BIC value is selected and its corresponding number of clusters $n_c$ are used for clustering the data.

\subsection{Kohonen Network for data clustering}
The DRBM-ClustNet uses the feature extracted data from DRBM and the BIC selected number of clusters $(n_c)$. The data clustering is performed using the Kohonen Network. The Kohonen Network (KN) is made up of input and output layer. The number of output layer neurons $(n_o)$ depends on the total number of clusters $(n_c)$ predicted using BIC. The weights between the input and the output layer are $W_k \in R^{n_i\times n_c}$, where $n_i$ is the number of input neurons and the weight matrix is randomly initialized. The weight is updated iteratively using the discriminant function considering Euclidean distance given by

\begin{equation}
    d(j) = \sqrt{\sum_{i=1}^{n_i}(\tilde{x}_i - w_{ij})} \:\:\:\:\:\:\:\:\: j \in 1\:to\: n_o.
\label{eq:eq17}
\end{equation}

\noindent
The winning neuron is the one, which has minimum separation with the input sample. The weight is updated iteratively considering the neighborhood function (i.e., the neurons that are within its vicinity) using

\begin{equation}
    h_{ci}(t) = \alpha(t)exp\left( \frac{-d_{ij}^{2}}{2\sigma(t)} \right),
\label{eq:eq18}
\end{equation}

\noindent
where $t$ is the iteration, $\alpha(t)$ is the learning rate for given iteration $t$ and $\sigma(t)$ is the spread of the data points in consideration and is given by
\begin{equation}
    \alpha(t) = \alpha_0exp\left( \frac{-t}{T_1} \right),
\label{eq:eq19}
\end{equation}

\begin{equation}
    \sigma(t) = \sigma_0exp\left( \frac{-t}{T_2} \right),
\label{eq:eq19}
\end{equation}

\noindent
where $T_1$ and $T_2$ are time constants.

The updating of the weights for the winning neuron and its neighboring neurons within its vicinity are computed as follows:

\begin{equation}
    \delta w_{ij} = h_{ci}(t)(y_i - w_{ij}).
\label{eq:eq20}
\end{equation}

DRBM-ClustNet uses this as a final layer for data clustering. To know the grouping of the data points to individual class label is evaluated using a performance measure. The clustering efficiency can be evaluated using the clustering accuracy ($\eta$)

\begin{equation}
    \eta = \frac{\sum_{i=1}^{n_c}a_l}{N},
\label{eq:eq21}
\end{equation}

\noindent
where $a_l$  is the correctly clustered samples and $N$ is the total number of samples in the data. The proposed DRBM-ClustNet is summarized in Algorithm 1.

\begin{algorithm}[H]
\caption*{\textbf{Algorithm 1} DRBM-ClustNet Algorithm}
\label{alg:TEC1}
\textbf{Input:} clustering dataset \textbf{X}=${\{\textbf{x}_s\in {\rm I\!R}^d\}_{s=1}^N}$, DRBM-ClustNet ($v_1,...,v_d,h_1,...,h_p$), $\theta_1$=$\{W^l\}_{l=1}^n$, $\theta_2$=\{$W_K$\}, learning rate ($\epsilon$, $\alpha$).\\
\textbf{Output:} cluster prediction $(n_c)$, clustering accuracy ($\eta$).\\
\For{\textbf{x}$_s$ in \textbf{X}}{
\textbf{v}$_{i}^{s}$\texttt{<-}\textbf{x}$_s$\\
\For{$l = 1$ to $n$}{
    Compute CD using (\ref{eq:eq9})-(\ref{eq:eq14}) to obtain $W^l$
   Perform feedforward pass using $W^l$ 
   Extract features 
   $\tilde{\textbf{x}}_{s}^{l}$ from layer \textbf{h}$^l$
}
\noindent Using extracted feature  $\tilde{\textbf{x}}_{s}^{l}$ predict $n_c$ (\ref{eq:eq16})
        
\noindent Perform clustering with $n_c$ on $\tilde{\textbf{x}}_{s}^{l}$ using (\ref{eq:eq17})-(\ref{eq:eq20})

\noindent Evaluate $\eta$ using (\ref{eq:eq21})
}
\textbf{return:} $n_c$, $\eta$. 
\end{algorithm}

\section{DRBM-ClustNet applied on synthetic dataset}
\begin{figure*}[h]
\centering
    \captionsetup{justification=centering}
\includegraphics[width=0.8\linewidth]{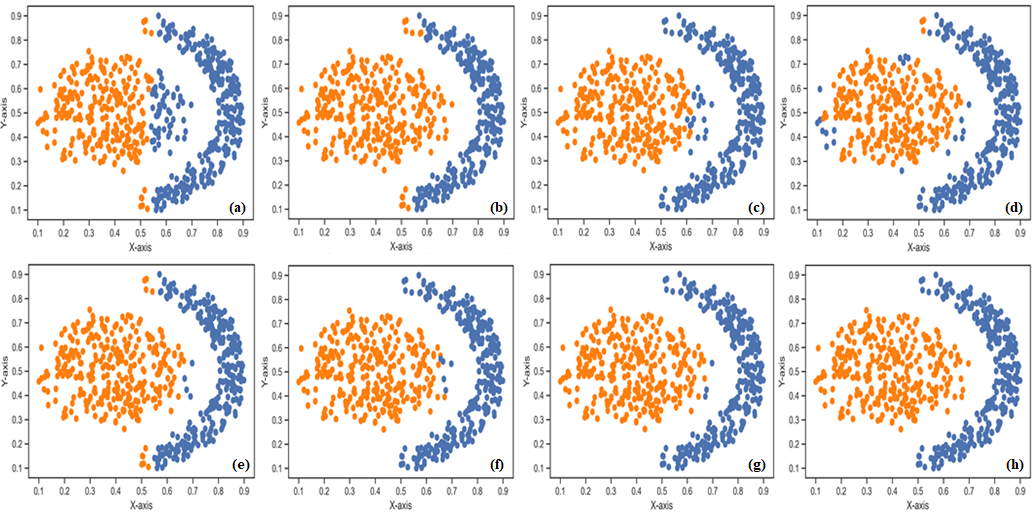}
    \caption{Clustering of flame pattern distribution using (a) k-means; (b) SOM; (c) EM; (d) DBSCAN; (e) US-ELM; (f) BELMKN; (g) Single Layer RBM; and (h) DRBM-ClustNet}
    \label{Flame}
\end{figure*}

\begin{figure*}[h]
    \centering
        \captionsetup{justification=centering}
    \includegraphics[width=0.8\linewidth]{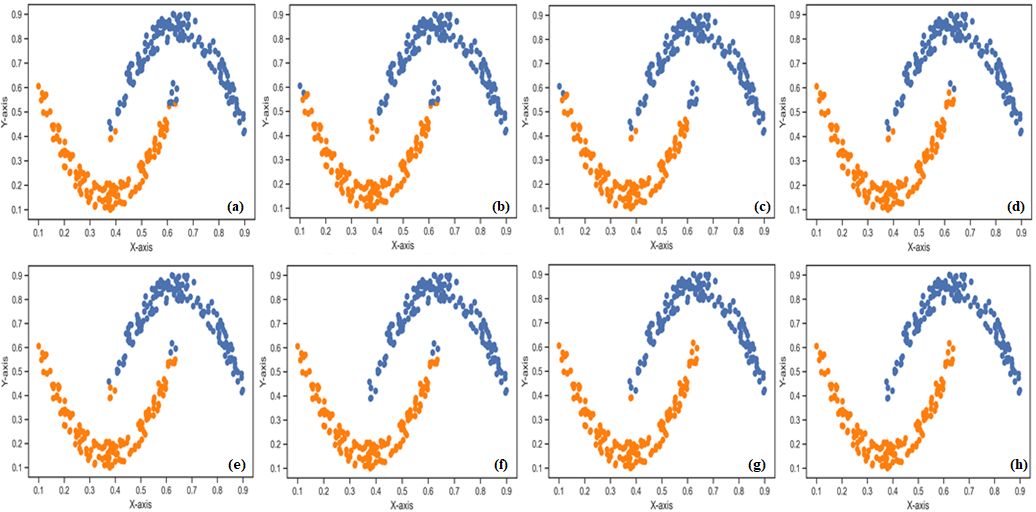}
    \caption{Clustering of moon pattern distribution using (a) k-means; (b) SOM; (c) EM; (d) DBSCAN; (e) US-ELM; (f) BELMKN; (g) Single Layer RBM; and (h) DRBM-ClustNet}
    \label{Moon}
\end{figure*}

In this section, we discuss the results of the proposed framework and other clustering algorithms for two synthetic datasets as shown in Figs. \ref{Flame} and \ref{Moon}. The results obtained are evaluated and benchmarked against the state-of-the-art clustering algorithms such as K-means \cite{n35}, SOM \cite{n36}, EM \cite{n37} and DBSCAN \cite{nw3}. Similarly, clustering is compared with feature learning algorithms like US-ELM \cite{n38}, BELMKN \cite{n39}, and Single Layer RBM \cite{n16}. 

The first dataset is of the flame distribution, which consists of 600 instances. It has two classes with 300 instances each. The second dataset is of the moon distribution, which consists of two classes (two half-moons) each of which has 150 samples. The flame distribution and the moon distribution datasets are non-linearly distributed and clustering them accurately is a challenge.

For each of the two datasets, DRBM-ClustNet is applied and compared with the aforementioned clustering techniques and their performance is analyzed. DRBM-ClustNet uses DRBM for feature extraction by setting the parameters such as the number of hidden layers and the number of hidden neurons in each layer. We can observe from Fig. \ref{Flame} that K-means, SOM, EM, DBSCAN and US-ELM have failed to cluster the dataset efficiently. However, BELMKN clusters the dataset efficiently as it uses ELM for feature learning and the Kohonen Network for clustering. The clustering accuracy using BELMKN for the flame distribution dataset is 98.5\%. The single layer RBM has few misclassification samples to capture the non-linearity in the dataset. It is overcome by stacking the multiple layers of RBM to form a DRBM, which is used in the proposed DRBM-ClustNet. The DRBM-ClustNet architecture for the flame distribution dataset consists of five hidden layers. For the proposed DRBM-ClustNet, we can observe that there is no misclassification for the flame distribution dataset, where the performance is better compared to the other clustering algorithms used in this paper.

Similarly, we can observe from Figs. \ref{Moon}(a)-(g) that for the moon distribution dataset, all the clustering algorithms fail to cluster the corners of the moon-shaped distribution. Fig. \ref{Moon}(h) shows the proposed method DRBM-ClustNet can cluster even the corners without any samples being misclassified.

By observing the two synthetic datasets used here, we can infer that the proposed DRBM-ClustNet which uses DRBM for feature extraction gives a better clustering accuracy in comparison to state-of-the-art clustering algorithms. Hence, we conclude that the DRBM-ClustNet can be used for efficient clustering of non-linearly distributed datasets.

\section{Results and Discussion}
This section discusses the results for the DRBM-ClustNet framework on fifteen different benchmark datasets from the UCI repository. Initially, we discuss the characteristics of the datasets shown in Table \ref{table:data_prop}. Next, the number of clusters are predicted using BIC with the feature extracted data obtained using three feature learning techniques, namely, ELM, Single Layer RBM and DRBM-ClustNet. The cluster prediction of these feature learning techniques is used for comparison against the BIC evaluated on the actual dataset. Finally, clustering performance of the DRBM-ClustNet algorithm is compared with other prominent clustering algorithms such as K-means \cite{n35}, SOM \cite{n36}, EM \cite{n37}, DBSCAN \cite{nw3}, USELM \cite{n38} and BELMKN \cite{n39}. The algorithms were executed on a Core i3 processor, 4 GB RAM, Python 2.7 and Windows 10 OS.

\subsection{Dataset Description}
In this research, we have applied the proposed DRBM-ClustNet on 15 benchmark datasets as shown in Table \ref{table:data_prop} (https://archive.ics.uci.edu/ml/index.php). We have adopted 15 widely used datasets of the UCI repository to compare with other prominent clustering methods. The number of samples/observations, the number of features/attributes and the actual number of clusters for every dataset used in this work is shown in Table \ref{table:data_prop}. The detailed description for the datasets are as follows:

\noindent
Dataset 1: The Balance data classifies a sample into one of the three classes, namely, balance scale tip to the right, tip to the left or to be balanced. It consists of four attributes, which includes the left weight, the left distance, the right weight, the right distance, and 625 samples.

\noindent
Dataset 2: The Cancer data categorizes a tumor either as benign or malignant. It has 30 attributes and 569 observations in total.

\noindent
Dataset 3: The Cancer-Int data classifies a breast tumor either as malignant or benign. It has 699 observations and 9 attributes.

\noindent
Dataset 4: The Credit data has two classes, namely, to grant a credit approval or not. It has 14 attributes with 690 observations.

\noindent
Dataset 5: The Dermatology data has different diagnosis details for the erythemato-squamous diseases in dermatology. It has 6 classes, 34 attributes and 366 observations.

\noindent
Dataset 6: The Diabetes data is to diagnostically predict whether a patient has diabetes based on certain diagnostic measurements. It has 768 observations and 8 attributes.

\noindent
Dataset 7: The E.Coli data has details of the proteins cellular localization sites. It contains 5 classes, 327 observations and 7 attributes.

\noindent
Dataset 8: The Glass data has details for the oxide content and refractive index for different glass types. It consists of 6 classes, 214 observations and 9 attributes.

\noindent
Dataset 9: The Heart data consists of details for the diagnosis of heart diseases. It consists of 2 classes, 270 observations and 13 attributes.

\noindent
Dataset 10: The Horse data has the details for the different types of horses. It consists of 3 classes, 364 observations and 26 attributes.

\noindent
Dataset 11: The Iris data classifies a sample into one of the three types of flower classes, namely, setosa or virginica or versicolor. It has 150 observations and 4 attributes, namely, width of petals, length of petals, the width of sepals and length of the sepals.

\noindent
Dataset 12: The SECOM data has signals from various sensors during the process of manufacturing semiconductors. There are 1567 observations and 590 features. The dataset consists of two classes, non-fail and fail, and is highly imbalanced. The missing values in attributes were imputed with zeroes.

\noindent
Dataset 13: The Thyroid data consists of three classes, namely, normal function, under functional and over-functional. There are 215 observations and 5 attributes.

\noindent
Dataset 14: The Vehicle data consists of different types of vehicles, namely, bus, Opel, van and Saab. There is a total of 846 observations and 8 attributes.

\noindent
Dataset 15: The Wine data lists different types of chemical analysis of wine. It consists of 3 classes, 178 observations and 13 attributes.

\begin{table}[H]
\centering
\captionsetup{justification=centering}
\caption{Properties of the Datasets}
\begin{tabular}{@{}lllll@{}}
\toprule
Sl. No & Dataset   & Number of  & Input       & Number of   \\
       & name      & samples    & Dimensions  & Cluster \\
\midrule
1     & Balance    & 625        & 4           & 3   \\
2     & Cancer     & 569        & 30          & 2   \\
3     & Cancer Int & 699        & 9           & 2   \\
4     & Credit     & 690        & 14          & 2   \\
5     & Dermatology& 366        & 34          & 6   \\
6     & Diabetes   & 768        & 8           & 2   \\
7     & E.Coli     & 327        & 7           & 5   \\
8     & Glass      & 214        & 9           & 6   \\
9     & Heart      & 270        & 13          & 2   \\
10    & Horse      & 364        & 26          & 3   \\
11    & Iris       & 150        & 4           & 3   \\
12    & SECOM      & 1567       & 590         & 2  \\
13    & Thyroid    & 215        & 5           & 3   \\
14    & Vehicle    & 846        & 18          & 4   \\
15    & Wine       & 178        & 13          & 3   \\
\bottomrule
\end{tabular}
\label{table:data_prop}
\end{table}

\subsection{Parameters for Algorithm}
All the datasets used in the paper are pre-processed which includes imputation, encoding and scaling. In the imputation process, all the missing values were imputed with a zero. All the categorical features were label encoded or one-hot encoded based on feature description. All distance calculation is based on the Euclidean distance. Distance-based techniques are sensitive to the scale of features. Therefore, to avoid bias towards high-value features, all the features were normalized between $0.1$ to $0.9$ using min-max scaling. This also ensures a faster convergence for neural network-based approaches. It is necessary to tune the hyper-parameters for effective learning. Since our hyper-parameters are in a well-defined range, the standard grid search approach \cite{grid_search} has been used to find an optimal set of hyper-parameters for the data clustering. A single RBM had one hidden layer with $50$ hidden units. The deep RBM has five hidden layers with $50$ hidden units in each of the first four layers and $10$ hidden units in the final layer. The maximum epochs and the learning rate are set to $50000$ and $0.1$, respectively. Furthermore, as all these algorithms are dependent on initial random points or weights, ten runs were carried out, and the average and the standard deviation were recorded.

\subsection{Assessment of the cluster prediction by DRBM-ClustNet}
The BIC is used initially to evaluate the number of clusters for a given dataset. The proposed DRBM-ClustNet uses DRBM for feature extraction on the above-mentioned dataset. The feature extracted dataset is used to predict the number of clusters with BIC. Furthermore, the obtained results is compared with the feature extracted data based on ELM, the single layer RBM for the number of clusters predicted by BIC, along with the BIC applied on the actual data as shown in Table \ref{table:BIC}.

\begin{table*}[]
\centering
\captionsetup{justification=centering}
\caption{Cluster Prediction using BIC for 15 datasets}
\begin{tabular}{@{}llllll@{}}
\toprule
Dataset   & Actual   & BIC on           & ELM with   & Single Layer       & DRBM      \\
          & Clusters & actual data      & BIC        & RBM BIC            & with BIC  \\
\midrule
Balance   & 3        & $5 \pm 0.04$     & $7\pm0.31$ & $5\pm0$       & $\textbf{3}\pm0$   \\
Cancer    & 2        & $3 \pm 0.47$     & $3\pm0.26$ & $3\pm0$       & $3\pm0$   \\
Cancerint & 2        & $\textbf{2}\pm 0.63$     & $3\pm0.37$ & $3\pm0$       & $3\pm0$   \\
Credit    & 2        & $\textbf{2}\pm 0.39$     & $5\pm0.31$ & $4\pm0$       & $6\pm0$   \\
Dermatology& 6       & $5 \pm 0.99$     & $\textbf{6}\pm0.87$ & $\textbf{6}\pm0.44$       & $\textbf{6}\pm0.39$   \\
Diabetes  & 2        & $\textbf{2}\pm 0.93$     & $3\pm0.79$ & $5\pm0.31$       & $5\pm0.22$   \\
E.Coli    & 5        & $4 \pm 0.67$     & $\textbf{5}\pm0.61$ & $\textbf{5}\pm0$       & $\textbf{5}\pm0$   \\
Glass     & 6        & $3 \pm 0.81$     & $\textbf{6}\pm0.73$ & $\textbf{6}\pm0$       & $\textbf{6}\pm0$   \\
Heart     & 2        & $\textbf{2} \pm 1.07$     & $3\pm1.18$ & $4\pm0.94$       & $4\pm0.81$   \\
Horse     & 3        & $2 \pm 0.75$     & $2\pm0.88$ & $2\pm0.18$       & $\textbf{3}\pm0.18$   \\
Iris      & 3        & $\textbf{3} \pm 0.31$     & $\textbf{3}\pm0.02$ & $\textbf{3}\pm0$       & $\textbf{3}\pm0$   \\
SECOM   & 2        & $3 \pm 0.15$     & $\textbf{2}\pm 0.54$ & $\textbf{2}\pm 0.44$       & $\textbf{2}\pm 0.22$   \\
Thyroid   & 3        & $\textbf{3} \pm 0.39$     & $\textbf{3}\pm0.31$ & $\textbf{3}\pm0$       & $\textbf{3}\pm0$   \\
Vehicle   & 4        & $3 \pm 1.25$     & $3\pm1.10$ & $3\pm0.98$       & $\textbf{4}\pm0.65$   \\
Wine      & 3        & $\textbf{3} \pm 0.49$     & $\textbf{3}\pm0.23$ & $\textbf{3}\pm0$       & $\textbf{3}\pm0$   \\
\bottomrule
\end{tabular}
\label{table:BIC}
\end{table*}

In Table \ref{table:BIC}, we observe a slight change in the actual number of clusters in the database when compared with the BIC predicted number of clusters for various feature learning techniques. From Table \ref{table:BIC}, we can observe that for the Cancer and Cancer-Int datasets, none of the feature extracted data with BIC were able to predict the actual number of clusters. Here, instead of two clusters, all the feature learning techniques with BIC predict three clusters. This kind of behaviour is observed in Cancer and Cancer-Int datasets as they are linearly separable. By applying feature learning on these datasets, we are introducing additional non-linearity. As a result, feature learning techniques like ELM, Single Layer RBM and DRBM fail for these two datasets. In contrast, for the Dermatology, Iris, Thyroid and Wine datasets, the methods predicted the exact number of clusters. Furthermore, the Glass and SECOM datasets are highly non-linear, hence, we get the actual number of clusters by applying the feature learning techniques on these two datasets. For the Balance, Horse and Vehicle datasets, the USELM, BELMKN and single layer RBM feature learning data on BIC fail to accurately predict the number of clusters. Only the DRBM feature learning method accurately predicts the exact number of clusters as three, three and four for Balance, Horse and Vehicle datasets, respectively. This is observed, as these three datasets are highly non-linear and the non-linearity in the datasets can be captured by stacking many layers of RBM. As a result, we can conclude that the DRBM-BIC is a good approach for the prediction of the number of clusters for non-linearly distributed data.

\begin{figure*}[h]
    \centering
        \captionsetup{justification=centering}
    \includegraphics[height=6.5cm, width=15.5cm]{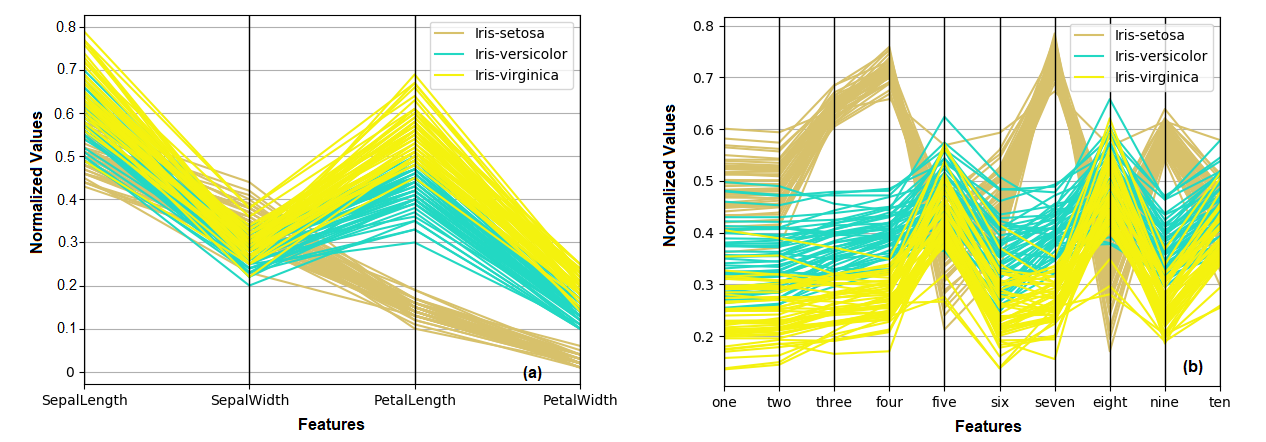}
    \caption{Parallel coordinates plot for (a) Actual Iris dataset, (b) Iris dataset after DRBM feature extraction}
    \label{Iris}
\end{figure*}

\begin{figure*}[h]
    \centering
        \captionsetup{justification=centering}
    \includegraphics[height=6.5cm, width=15cm]{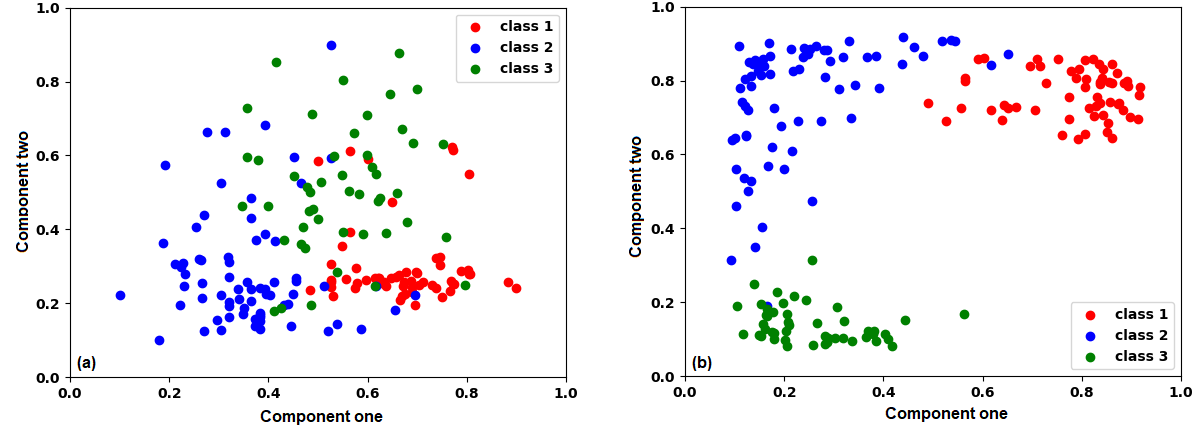}
    \caption{Scatter plot for (a) Actual Wine dataset, (b) Wine dataset after DRBM feature extraction}
    \label{Wine}
\end{figure*}

\begin{figure*}[h]
    \centering
        \captionsetup{justification=centering}
    \includegraphics[height=6cm, width=15cm]{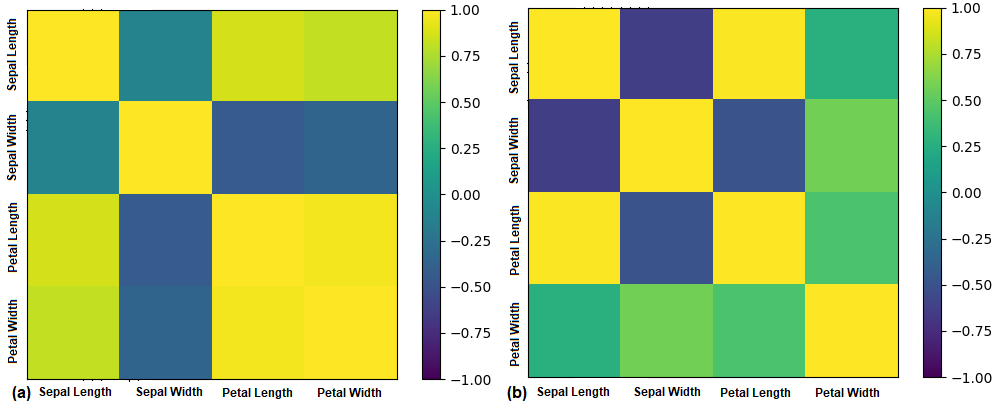}
    \caption{Correlation matrix plot for (a) Original Iris dataset, (b) Iris dataset after RBM feature extraction}
    \label{corr_matrix}
\end{figure*}

\subsection{Data visualization for DRBM}
To analyze the behaviour of data distribution in some of the scenarios where DRBM can pick the exact clusters, it is possible to visualize the data from high-dimension to low-dimension and vice versa. This is analyzed with and without applying DRBM on the dataset using a parallel coordinates plot, a scatter plot and a correlation matrix \cite{n42}.

The architecture used for DRBM feature learning has 10 hidden neurons in the last layer. The Iris dataset has four features and the DRBM feature learning, obtained in the last layer, contains 10 dimensions. Similarly, the Wine dataset has 13 features reduced to 10 features using 10 hidden neurons in the last layer of the DRBM architecture. The parallel coordinates and scatter plots for the Iris and Wine datasets without DRBM (actual data) and with DRBM feature extracted data are shown in Figs. \ref{Iris} and \ref{Wine}. From Fig. \ref{Iris}(a), for the Iris dataset, we can observe that there is an overlap between the two classes (class 2 and class 3). However, after applying DRBM feature learning, in Fig. \ref{Iris}(b), we can observe that the overlap is reduced with a clear separation between these classes. For the Wine dataset, we can observe from Figs. \ref{Wine}(a) and \ref{Wine}(b) that there is a significant difference between the two scatter plots. Because of the non-linear dimensionality reduction from 13 to 10 dimensions in the DRBM feature extraction, linear separability is introduced to the dataset. Hence, by applying DRBM feature extraction, the number of clusters is well-predicted using BIC as shown in Table \ref{table:BIC}.

Most machine learning algorithms involving sigmoidal activation functions or logistic regression functions, often exhibit poor performance if they have highly correlated input variables or feature vectors in the data. Fig. \ref{corr_matrix}(a) shows the correlation matrix for the actual Iris dataset. From this figure, we can observe that the percentage of feature vector pairs having a high correlation (in the range 0.75-1.0) is 6 out of 12 i.e. 50\%. In the case of DRBM feature leaning, as shown in Fig. \ref{corr_matrix}(b), we can observe that the percentage of attribute pairs having a high correlation (0.75-1.0) is 2 out of 12, i.e., 16.67\%. Hence, by applying DRBM feature extraction to the Iris dataset, we can significantly reduce the correlation between the feature vectors, which in turn results in high clustering accuracy.

Overall, from the above analysis, we can conclude that by decreasing the dimension, DRBM performs well for five datasets (Dermatology, Horse, SECOM, Vehicle and Wine). Hence, DRBM with BIC performs well for 10 out of the 15 datasets, i.e., in 66.6\% of the cases, whereas BIC applied on the actual dataset performs well for 7 out of 15 datasets, i.e., for 46.6\% of the datasets. Both ELM and the single layer RBM perform well for 7 out of 15 datasets, i.e., in 46.6\% of the datasets. Another observation from Table \ref{table:BIC} is that DRBM with BIC predicts cluster consistently as the standard deviation is almost zero for most of the datasets whereas a large variation is present for other techniques. Hence, DRBM works effectively and efficiently for the non-linearly distributed datasets.

\subsection{Assessment of the cluster accuracy of DRBM-ClustNet}
The proposed DRBM-ClustNet is applied to the above-mentioned 15 benchmark UCI machine learning repository datasets \cite{n40}. In the DRBM-ClustNet, the feature extracted data from the DRBM network and the cluster prediction output from BIC are given for clustering using the Kohonen Network. This approach shows the best performance when compared to the traditional clustering algorithms such as K-means, EM, SOM, DBSCAN and feature learning based clustering techniques such as USELM, BELMKN, and single layer RBM.

The K-means algorithm involves random initialization of centroids and the cluster centers are iteratively updated until it converges. The main drawback of the k-means is the random choice of initial clusters. In K-means, we assume that each attribute has the same weight, i.e., all attributes are assumed to have the same contribution towards clustering. In addition, it gets stuck in local optima, and we can observe from Table \ref{table:clust_acc} that the performance of the K-means algorithm is the worst of all the approaches. Unlike K-means, SOM gives weights for each attribute by using the neighbourhood concept and iteratively computes the cluster centers. The main drawback of SOM is ineffective in clustering non-linearly separable datasets as the hidden layers are absent. The EM algorithm iteratively computes the cluster centers using the maximum likelihood approach. The performance of EM is better than SOM and K-means as it is a soft clustering algorithm. The DBSCAN finds core samples of high density and expands clusters from them. However, as the quality of DBSCAN also depends on the distance measure used in the function region, it becomes challenging at higher dimensions to find appropriate separation or clusters.

\begin{table*}[]
\centering
\captionsetup{justification=centering}
\caption{Clustering accuracy percentage with standard deviation and ranking (in parenthesis below) of various techniques on each dataset}
\begin{tabular}{@{}cccccccccc@{}}
\toprule
Dataset     & K-means         & SOM             & EM         & USELM         & BELMKN        & DBSCAN     & Single RBM      & DRBM-ClustNet         \\
\midrule
Balance     & $51.4 \pm 0.9$  & $52.5\pm0.2$  & $53.0\pm0.6$ & $45.9\pm3.3$  & $47.1\pm3.4$  & $52.5 \pm 0.5$ & $68.1 \pm 0.16$    & $\textbf{70.8} \pm 0.06$    \\
            & (6)             & (5)           & (3)          & (8)           & (7)           & (4)        & (2)             & (1)               \\
Cancer      & $83.2 \pm 1.9$  & $86.0\pm1.0$  & $91.2\pm1.6$ & $83.8\pm1.6$  & $86.7\pm2.0$  & $88.8 \pm 0.8$ & $\textbf{95.0} \pm 0.2$ & $93.4 \pm 0.2$ \\
            & (8)             & (6)           & (3)          & (7)           & (5)           & (4)        & (1)             & (2)               \\
Cancerint   & $95.8 \pm 1.1$  & $94.2\pm0.4$  & $93.6\pm1.6$ & $92.0\pm1.5$  & $92.5\pm1.8$  & $91.9 \pm 0.9$ & $96.1 \pm 0.7$      & $\textbf{97.0} \pm 0.41$        \\
            & (3)             & (4)           & (5)          & (7)           & (6)           & (8)        & (2)             & (1)               \\
Credit      & $52.2 \pm 2.8$  & $54.8\pm1.2$  & $51.4\pm1.8$ & $58.5\pm4.1$  & $61.5\pm3.8$  & $67.1 \pm 1.3$ & $85.3 \pm 0.13$ & $\textbf{86.5} \pm 0.26$ \\
            & (7)             & (6)           & (8)          & (5)           & (4)           & (3)        & (2)             & (1)    \\
Dermatology & $24.2 \pm 3.9$  & $29.8\pm1.5$  & $67.7\pm0.4$ & $71.0\pm3.9$  & $76.6\pm3.6$  & $63.2 \pm 0.9$ & $92.8 \pm 0.4$      & $\textbf{95.5} \pm 0.28$    \\
            & (8)             & (7)           & (5)          & (4)           & (3)           & (6)        & (2)             & (1)               \\
Diabetes    & $63.6 \pm 1.5$  & $64.8\pm0.8$  & $52.4\pm0.5$ & $65.0\pm1.9$  & $67.9\pm2.4$  & $61.7 \pm 0.8$ & $69.2 \pm 0.23$    & $\textbf{71.0} \pm 0.25$    \\
            & (6)             & (5)           & (8)          & (4)           & (3)           & (7)        & (2)             & (1)               \\
E.Coli      & $53.8 \pm 5.7$  & $61.0\pm3.8$  & $82.0\pm0.5$ & $80.1\pm2.7$  & $80.7\pm1.8$  & $52.8 \pm 4.2$ & $80.9 \pm 1.2$  & $\textbf{83.1} \pm 0.19$  \\
            & (7)             & (6)           & (2)          & (5)           & (4)           & (8)        & (3)             & (1)               \\
Glass       & $52.1 \pm 3.8$  & $53.8\pm2.9$  & $47.7\pm0.4$ & $40.2\pm4.1$  & $41.5\pm3.5$  & $39.8 \pm 3.7$ & $54.4 \pm 0.21$  & $\textbf{57.6} \pm 0.26$ \\
            & (4)             & (3)           & (5)          & (7)           & (6)           & (8)        & (2)             & (1)               \\
Heart       & $58.2 \pm 3.4$  & $58.8\pm2.2$  & $52.6\pm1.6$ & $66.6\pm2.2$  & $71.4\pm2.9$  & $58.3 \pm 2.0$ & $79.2 \pm 0.27$   & $\textbf{82.4} \pm 0.24$ \\
            & (7)             & (5)           & (8)          & (4)           & (3)           & (6)        & (2)             & (1)   \\
Horse       & $49.7 \pm 4.4$  & $48.3\pm2.3$  & $43.4\pm1.9$ & $58.6\pm4.8$  & $60.9\pm4.1$  & $47.6 \pm 3.1$ & $62.0 \pm 0.56$   & $\textbf{65.3} \pm 0.26$ \\
            & (5)             & (6)           & (8)          & (4)           & (3)           & (7)        & (2)             & (1)   \\
Iris        & $88.6 \pm 0$  & $81.3\pm0.8$  & $90.0\pm0.3$ & $91.1\pm5.3$  & $92.3\pm4.8$  & $77.2 \pm 1.8$ & $91.5 \pm 0.33$  & $\textbf{93.8} \pm 0.33$ \\
            & (6)             & (7)           & (5)          & (4)           & (2)           & (8)        & (3)             & (1)     \\
SECOM & $62.2 \pm 2.2$ & $60.6\pm0.3$  & $66.1\pm0.7$  & $65.9\pm1.2$  & $69.4\pm2.3$  & $61.6 \pm 0.9$ & $73.1 \pm 0.40$ & $\textbf{73.5} \pm 0.28$  \\
            & (6)             & (8)           & (4)          & (5)           & (3)           & (7)        & (2)             & (1)      \\
Thyroid     & $85.8 \pm 2.2$  & $86.2\pm0.3$  & $90.1\pm0.7$ & $86.5\pm1.2$  & $87.6\pm2.3$  & $85.6 \pm 0.9$ & $\textbf{94.1} \pm 0.09$  & $92.0 \pm 0.09$  \\
            & (7)             & (6)           & (3)          & (5)           & (4)           & (8)        & (1)             & (2)   \\
Vehicle     & $44.0 \pm 2.2$  & $44.0\pm1.5$  & $45.0\pm1.8$ & $40.5\pm3.8$  & $42.0\pm2.2$  & $43.3 \pm 1.5$ & $46.0 \pm 0.62$  & $\textbf{48.4} \pm 0.22$  \\
            & (4)             & (4)           & (3)          & (8)           & (7)           & (6)        & (2)             & (1)   \\
Wine        & $70.0 \pm 3.4$  & $75.0\pm0.0$  & $90.4\pm0.3$ & $91.9\pm1.1$  & $95.5\pm1.7$  & $74.8 \pm 0.4$ & $95.8 \pm 0.08$ & $\textbf{97.2} \pm 0.08$  \\
            & (6)             & (8)           & (4)          & (5)           & (3)           & (7)        & (2)             & (1)      \\
\bottomrule
\end{tabular}
\label{table:clust_acc}
\end{table*}

\begin{table*}[]
\centering
\captionsetup{justification=centering}
\caption{Average clustering accuracy and general ranking of the techniques for all datasets}
\begin{tabular}{@{}cccccccccc@{}}
\toprule
Dataset     & K-means         & SOM             & EM         & USELM         & BELMKN        & DBSCAN     & Single RBM      & DRBM-ClustNet         \\
\midrule
Average & $62.32$  & $63.41$  & $67.77$ & $69.16$  & $71.55$  & $64.41$ & $78.92$    & $\textbf{80.59}$    \\
Rank        & 8   & 7   & 5  & 4 & 3  & 6  & 2  & 1 \\
\bottomrule
\end{tabular}
\label{table:avg_clust_acc}
\end{table*}

\begin{table*}[]
\centering
\captionsetup{justification=centering}
\caption{The sum of ranking of the techniques and general ranking based on total ranking}
\begin{tabular}{@{}cccccccccc@{}}
\toprule
Dataset     & K-means         & SOM             & EM         & USELM         & BELMKN        & DBSCAN     & Single RBM      & DRBM-ClustNet         \\
\midrule
Average & $92$  & $84$  & $75$ & $81$  & $63$  & $97$ & $30$ & $\textbf{17}$ \\
Rank        & 7   & 6   & 4  & 5 & 3  & 8  & 2  & 1  \\
\bottomrule
\end{tabular}
\label{table:rnk_clust_acc}
\end{table*}

The USELM and BELMKN perform better in comparison to K-means, DBSCAN, SOM and EM. USELM and BELMKN use ELM with K-means and Kohonen Network as the clustering algorithms, respectively. As BELMKN uses the Kohonen Network, this clustering algorithm overcomes the stated drawbacks of K-means. As both USELM and BELMKN use ELM for feature learning, they suffer from the drawbacks of ELM. ELM is a single hidden layer neural network where the weights between the input layer and hidden layer are randomly initialized and the weights between the hidden layer and the output layer are computed using a closed-form solution. This sometimes turns out to be a major drawback as it increases the amount of randomness in the network and hence the clustering results obtained through ELM feature learning are not consistent which may result in either underfitting or overfitting. From Table \ref{table:clust_acc}, we can observe that the standard deviation for USELM and BELMKN is greater than 1 for all the datasets. For the datasets that are highly non-linear like Balance, Credit, Dermatology, Glass and Heart, the standard deviation is very high compared to the other datasets which indicate that the clustering results are not consistent for ELM. Hence, for USELM and BELMKN we need to run the program several trials to obtain better accuracy. Another disadvantage of ELM is that it sometimes overfits the data. As a result, more samples are drawn into one of the clusters that dominate the data and the remaining clusters suffer from the sparsity of samples. Hence, we can observe from Table \ref{table:clust_acc} that for datasets like Balance, Dermatology and Thyroid, the clustering accuracies of the ELM based algorithms USELM and BELMKN are lower than the ones obtained by the proposed DRBM-ClustNet. 

DRBM-ClustNet is modelled as bipartite graphical model to exploit the generative nature of data and to estimate better probability distributions. Each hidden neuron in a DRBM-ClustNet is a stochastic processing unit, which learns a probability distribution over the inputs. The hidden units in a DRBM-ClustNet capture higher-order correlations between the inputs. In DRBM-ClustNet, the convergence to the minima is achieved in two steps, namely based on the positive contrastive divergence and the negative contrastive divergence, which is accomplished by Gibbs sampling \cite{n18}. Hence, the convergence to a global minimum is better guaranteed when compared to ELM-based feature learning. As a result, the clustering accuracy for the datasets using DRBM-ClustNet is higher when compared to USELM and BELMKN. In addition, there are higher accuracies obtained by DRBM-ClustNet for the non-linear datasets like Balance, Credit, Dermatology, Glass and Heart. As DRBM-ClustNet is less immune to the non-linearity in the dataset, it does not overfit the data. 

The RBM network with one hidden layer is not sufficient to capture the non-linearity of a dataset even with an increase in the number of hidden neurons in that layer. This is evident for almost all the datasets from Table \ref{table:clust_acc}. The proposed DRBM-ClustNet overcomes this problem during the processing of input data. Initially it has one hidden layer but the output of this layer is taken to the next hidden layer and a stack of RBM layers are built up producing a better generative model. The DRBM-ClustNet has the potential to learn internal representations of the data that are increasingly complex. As a result, DRBM-ClustNet performs better in terms of accuracy when compared to the RBM with one hidden layer and also in comparison with the other clustering algorithms used in this study.

Furthermore, Table \ref{table:clust_acc} shows that the clustering results using single layer RBM and DRBM-ClustNet are consistent for every dataset. The standard deviation is less than about 0.7 and 0.41 in all cases for the single layer RBM and DRBM-ClustNet, respectively. The proposed approach also works for clustering of high dimensional and highly imbalanced data. This can be observed by looking at the results in the Table for the SECOM data which has 590 attributes and a minority class of $6\%$. Here, it can be stated that the DRBM-ClustNet not only outperforms the other algorithms in terms of accuracy but also does well when a minority class is in the data. 

From Table \ref{table:clust_acc}, for the Thyroid dataset, average clustering accuracies with RBM with one hidden layer and DRBM-ClustNet is $94.1\%$ and $92.0\%$, respectively. The same kind of results are also observed for the Cancer dataset. There is an exception here in that the feature extraction with RBM with one hidden layer produces a higher clustering accuracy when compared to the DRBM-ClustNet. This occurs because the Thyroid dataset is a more linearly separable dataset. Increasing the number of hidden layers for a simple dataset may result in overfitting. As a result, a drop in the accuracy of the DRBM-ClustNet is observed. Overall, from Table \ref{table:clust_acc} it is evident that for 13 out of 15 datasets (i.e. $86.6\%$) the DRBM-ClustNet performs best in terms of accuracy when compared to all other clustering approaches.

Table \ref{table:avg_clust_acc} shows the average clustering accuracy for each of the clustering approaches applied on the 15 benchmark datasets. Here, we can infer that the DRBM-ClustNet has the maximum average clustering accuracy. The single-layer RBM ranks second, then BELMKN, followed by USELM. Amongst the traditional clustering approaches, EM fares better than DBSCAN, SOM and K-means. Table \ref{table:rnk_clust_acc} represents the sum of the ranks for all the datasets and clustering approaches shown in Table \ref{table:clust_acc}. The ranking based on the sum of ranks indicates that the proposed DRBM-ClustNet outperforms the other clustering approaches. The Single RBM, BELMKN, EM, USELM, SOM, K-means and DBSCAN follow DRBM-ClustNet in ranking order.

\subsection{Assessment of the clustering on image datasets}
In this section, the proposed algorithm is compared with state-of-the-art algorithms based on RBM, namely Graph  RBM, mixed GraphDBN (mGraphDBN) and full GraphRBM-based DBN (fGraphDBN) \mbox{\cite{n43}}. The performance of the DRBM-ClustNet algorithm is assessed on three publicly available image datasets, namely, COIL-20, the Extended Yale database B (YaleB) and MNIST. All the observations in COIL-20 and YaleB are used for experimentation. In the case of MNIST data, 60000 training dataset observations are used in the experimental study as in  \mbox{\cite{n43}}. The Whitening Principal Component Analysis is applied for reducing the dimension of all the three image datasets to 400 as in \mbox{\cite{n43}}. The setting of DRBM-ClustNet remains the same as in the previous section. The obtained results are also compared with the normalized mutual information (NMI) as performance metric  \mbox{\cite{n43}} \mbox{\cite{n44}} and tabulated in Table \mbox{\ref{table:image_data}}. The populated values are from ten runs and indicate the mean with its standard deviation. The performance of other algorithms in  \mbox{\cite{n43}} is listed for reference in the Table.

From Table \ref{table:image_data} it can be observed that DRBM-ClustNet performs best for the MNIST and YaleB datasets. In the case of COIL-20, DRBM-ClustNet performs better than GraphRBM and mGraphDBN but shows a fractionally lower performance than fgraphDBN. This performance could be attributed to better subspace clustering  by DRBM-ClustNet in the final stage, which uses the Kohonen network. Furthermore, it is also observed that the average number of clusters predicted by the second stage of DRBM-ClustNet for COIL-20, YaleB, and MNIST datasets are $19\pm0.87$, $38\pm0.61$, and $10\pm0.21$, respectively. The actual number of clusters for COIL-20, YaleB, and MNIST dataset are $20$, $38$, and $10$, respectively. This indicates the ability of DRBM-ClustNet to accurately pick the number of clusters without having to define them explicitly like other methods. Overall, these results demonstrate the superior performance of the DRBM-ClustNet framework in comparison to other state-of-the-art RBM algorithms for various non-linear image datasets.

\begin{table}[]
\centering
\captionsetup{justification=centering}
\caption{Clustering performance (NMI value) on the COIL-20, YaleB and MNIST datasets}
\begin{tabular}{@{}ccccc@{}}
\toprule
Algorithms & COIL-20 & YaleB & MNIST \\
\midrule
SSC$^1$ & $80.72 \pm 0.88$ & $63.42 \pm 0.79$ & $68.35 \pm 0.02$ \\
LSC$^1$ & $65.25 \pm 2.66$ & $41.96 \pm 0.98$ & $74.06 \pm 2.75$ \\
SCC$^1$ & $88.23 \pm 1.37$ & $34.39 \pm 0.61$ & $57.00 \pm 0.04$ \\
LRR$^1$ & $65.60 \pm 1.02$ & $10.19 \pm 0.18$ & $46.10 \pm 0.01$ \\
LRSC$^1$ & $67.55 \pm 1.52$ & $14.94 \pm 0.39$ & $50.96 \pm 0.00$ \\
LSR1$^1$ & $72.33 \pm 0.88$ & $66.93 \pm 1.61$ & $45.73 \pm 0.01$ \\
LSR2$^1$ & $72.18 \pm 1.08$ & $67.04 \pm 1.62$ & $45.72 \pm 0.02$ \\
GGMM$^1$ & $79.53 \pm 1.26$ & $12.37 \pm 0.31$ & $62.34 \pm 1.69$ \\
GSC$^1$ & $78.02 \pm 3.58$ & $26.55 \pm 1.00$ & $70.57 \pm 2.95$ \\
GNMF$^1$ & $72.00 \pm 0.08$ & $49.10 \pm 1.37$ & $72.00 \pm 1.01$ \\
\hline
gSAE$^1$ & $57.10 \pm 2.88$ & $16.53 \pm 1.40$ & $17.12 \pm 0.54$ \\
RBM$^1$ & $73.81 \pm 1.79$ & $18.53 \pm 1.64$ & $53.50 \pm 1.61$ \\
DBN$^1$ & $76.32 \pm 2.63$ & $21.77 \pm 1.57$ & $52.79 \pm 2.44$ \\
DAE$^1$ & $75.29 \pm 1.73$ & $18.37 \pm 2.00$ & $50.57 \pm 1.31$ \\
SDAE$^1$ & $77.72 \pm 0.51$ & $20.99 \pm 1.84$ & $52.81 \pm 1.61$ \\
\hline
GraphRBM$^1$ & $94.46 \pm 2.14$ & $73.03 \pm 1.32$ & $86.66 \pm 1.23$ \\
mGraphRBM$^1$ & $95.19 \pm 1.98$ & $73.36 \pm 0.93$ & $88.73 \pm 1.49$ \\
fGraphRBM$^1$ & $\textbf{95.54} \pm 1.43$ & $75.18 \pm 1.29$ & $89.97 \pm 0.98$ \\
\hline
DRBM-ClustNet & $95.06 \pm 1.56$ & $\textbf{76.06} \pm 1.11$ & $\textbf{90.11} \pm 1.03$ \\
\bottomrule
$^1$Reproduced from \cite{n43}
\end{tabular}
\label{table:image_data}
\end{table}

\begin{figure*}[h]
    \centering
        \captionsetup{justification=centering}
    \includegraphics[height=5cm, width=17.5cm]{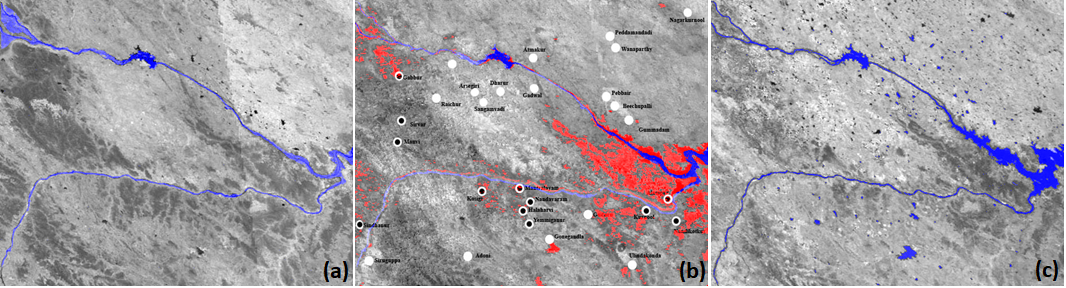}
    \caption{Assessment flooded region: (a) before flood, (b) during flood, (c) after flood}
    \label{flood_assess}
\end{figure*}

\noindent {\textbf{\textit{Assessment of the flooded region using satellite imagery:}}
The DRBM-ClustNet is further evaluated for clustering actual satellite images of the flood-prone region. This application would aid in the quick identification of the extent of flood-prone regions. Here, three MODIS satellite images of various stages, namely, before the flood, during the flood and after the flood are used as shown in Fig. \mbox{\ref{flood_assess}}. The dimension of each of the three images is 482 x 627 pixels. Details of the MODIS images used in this study are discussed in \mbox{\cite{n45}}. All three images were processed using DRBM-ClustNet for spectral clustering to assess the extent of water (flood) cover. The extracted flood regions across all three images are highlighted in the figure with blue (actual river) and red (flooded region) colors.

MODIS satellite images are of low spatial resolution (250 m), hence the study prominently focuses on city granularity for the current application. Fig. \mbox{\ref{flood_assess}} (b) shows results of the proposed approach during a flood timeline wherein white dots signify cities that were not flooded while the flooded cities are indicated with white dots with black dots inside. Here, it can be observed that the DRBM-ClustNet is able to correctly identify flooded regions for 12 out of 15 cities with just one false positive. Other clustering approaches such as K-means and SOM are able to identify only 6 and 8 cities with false positives being 4 and 5 respectively. These numbers reflect the inability to identify the majority of flooded regions accurately.  In the case of before the flood and after the flood, K-means clustering accuracies were $83.87\%$ and $85.21\%$ and SOM clustering accuracies were $87.54\%$ and $89.05\%$, respectively. However, the clustering accuracy of DRBM-ClustNet for before the flood and after the flood were $97.3\%$ and $98\%$, respectively.  Overall, the DRBM-ClustNet extracted images show a better performance than the other considered approaches. Therefore, DRBM-ClustNet can be further explored for other practical image clustering tasks.}

\section{Conclusion}
In this paper, DRBM-ClustNet is proposed, a Bayesian Deep Restricted Boltzmann-Kohonen Network based on clustering. DRBM-ClustNet uses three stages to maximize the clustering efficiency, particularly for non-linearly separable datasets. The first stage uses the DRBM, a generative model built by stacking multiple layers of RBM for feature learning. The second stage uses BIC to calculate the number of actual clusters in the dataset. The final stage uses the Kohonen Network to cluster the feature extracted data from DRBM using BIC predicted clusters. The parameters of the DRBM-ClustNet are set by the grid search approach, i.e., to determine the number of hidden layers and the number of hidden neurons to obtain a better clustering accuracy.

The data clustering task was successfully accomplished with DRBM-ClustNet on two synthetic datasets, fifteen UCI repository benchmark data and four image datasets.The results demonstrate that the proposed DRBM-ClustNet approach outperforms the other clustering techniques used in experiments. DRBM-ClustNet is also able to accurately predict the number of clusters for various datasets, giving it a further edge over other techniques. The obtained results show that DRBM-ClustNet is a very consistent, efficient and reliable algorithm for clustering of non-linearly separable datasets.

\section*{Appendix}
\noindent Recall, in DRBM-ClustNet framework the stage 1 consists of DRBM, here the joint probability of the visible and hidden units is given by

\begin{equation}
    P(v,h) = \frac{e^{-E(v,h)}}{\sum_v\sum_he^{-E(v,h)}},
\label{eq:eq23}
\end{equation}

\noindent where

\begin{equation}
    E(v,h) = -\sum_{i,j}v_{i}h_jW_{ij} - \sum_{i}v_ib_i - \sum_{j}h_jc_j,
\label{eq:eq24}
\end{equation}

\begin{equation}
    P(v,h) = \frac{1}{Z}e^{-E(v,h)},
\label{eq:eq25}
\end{equation}

\noindent where Z is called the partition function which is given by

\begin{equation}
    Z = \sum_v\sum_he^{-E(v,h)}.
\label{eq:eq26}
\end{equation}

The partition function $Z$ is intractable and the joint probability $P(v,h)$ is also intractable.
Hence, we derive the conditional probability distribution $P(h| v)$ and $P(v|h)$  from the joint probability distribution $P(v,h)$ which is easy to compute and to sample from.

\noindent Equation (\ref{eq:eq24}) can be expressed in matrix form as

\begin{equation}
    E(v,h) = -b^Tv - c^Th - v^TWh,
\label{eq:eq27}
\end{equation}

\begin{equation}
    P(v,h) = \frac{1}{Z}e^{(b^Tv + c^Th + v^TWh)},
\label{eq:eq28}
\end{equation}

\begin{equation}
    P(h|v) = \frac{P(h,v)}{P(v)},
\label{eq:eq29}
\end{equation}

\begin{equation}
    P(h|v) = \frac{1}{P(v)}\frac{1}{Z}e^{(b^Tv + c^Th + v^TWh)}.
\label{eq:eq30}
\end{equation}

\noindent The derivative $Z'$ using (\ref{eq:eq28}) can be expressed in terms of $Z$ as

\begin{equation}
    Z' = \frac{e^{(b^Tv)}}{P(v)Z}.
\label{eq:eq31}
\end{equation}

\noindent Substituting (\ref{eq:eq31}) in  (\ref{eq:eq30}), we get

\begin{equation}
    P(h|v) = \frac{1}{Z'}\prod_{j=1}^{p}e^{(c_jh_j + v^TW_{ij}h_j)}.
\label{eq:eq32}
\end{equation}

\noindent Using Bayes theorem

\begin{equation}
    P(h_j=1|v) = \frac{P(h_j=1,v)}{P(h_j=0,v)+P(h_j=1,v)}.
\label{eq:eq33}
\end{equation}

\noindent Substitute $h_j=1$, in (\ref{eq:eq32}) to get

\begin{equation}
    P(h_j=1|v) = sigmoid(c_j + v^TW_{ij}).
\label{eq:eq34}
\end{equation}

\noindent Similarly for $P(v|h)$ we obtain

\begin{equation}
    P(v_i=1|h) = sigmoid(b_i + W_{ij}h).
\label{eq:eq35}
\end{equation}

\noindent Therefore 

\begin{equation}
    P(h|v) = \prod_{j=1}^{p}sigmoid(c_j + v^TW_{ij}),
\label{eq:eq36}
\end{equation}

\noindent and

\begin{equation}
    P(v|h) = \prod_{i=1}^{d}sigmoid(b_i + W_{ij}h).
\label{eq:eq37}
\end{equation}

Figs. \ref{Wine}(a) and \ref{Wine}(b) show the cluster separation for the Wine dataset before applying DRBM and after applying DRBM respectively, using the conditional probability distributions as mentioned above.

We can observe that in Fig. \ref{Wine}(a), the clusters are overlapping before applying DRBM and the clusters are not so obvious, making it difficult for BIC to predict the number of clusters and also for SOM to calculate the cluster centroids. In Fig. \ref{Wine}(b), the clusters are well separated which makes it easy for BIC and SOM to predict the number of clusters and calculate the cluster centroids respectively. The same behavior is also observed for the other datasets used for comparing the clustering accuracy.

BIC uses the maximum likelihood parameter to determine the number of cluster which can also be expressed in terms of Mean Squared Error (MSS) as follows

\begin{equation}
    BIC = ln(N)k - 2 ln(\Tilde{L}),
\label{eq:eq38}
\end{equation}

\noindent where $\Tilde{L}$ is the maximum likelihood function

Assuming model errors are independent and identically distributed, according to the normal distribution theory

\begin{equation}
    h_{ML} = \prod_{i=1}^{p}P(\frac{d_i}{p}),
\label{eq:eq39}
\end{equation}

\noindent where $h_{ML}$ is the maximum likelihood hypothesis, $p$ is the number of hidden layers and $d_i$ is the output from DRBM

\begin{equation}
    h_{ML} = \prod_{i=1}^{p}\frac{e^{-\frac{(d_i-\mu)^2}{2\sigma^2}}}{\sqrt{2\pi\sigma^2}},
\label{eq:eq40}
\end{equation}

\noindent where $d_i=f(x_i)+e_i$, $\mu$ and $\sigma$ are the mean and variance of normal distribution, $f(x_i)$ is a function of $x_i$ and $e_i$ is random variable representing noise. Hence,

\begin{equation}
    h_{ML} = \prod_{i=1}^{p}\frac{e^{-\frac{(d_i-f(x_i))^2}{2\sigma^2}}}{\sqrt{2\pi\sigma^2}},
\label{eq:eq41}
\end{equation}

\begin{equation}
    log(h_{ML}) = \sum_{i=1}^{p} ln(\frac{1}{\sqrt{2\pi\sigma^2}}) - \frac{(d_i-f(x_i))^2}{2\sigma^2}.
\label{eq:eq42}
\end{equation}

The first term in the above equation is independent of the function $f$ and it can be discarded, therefore

\begin{equation}
    log(h_{ML}) = \sum_{i=1}^{p} - \frac{(d_i-f(x_i))^2}{2\sigma^2},
\label{eq:eq43}
\end{equation}

\begin{equation}
    - log(h_{ML}) = - \sum_{i=1}^{p} - \frac{(d_i-f(x_i))^2}{2\sigma^2} = \frac{RSS}{n} = MSE.
\label{eq:eq44}
\end{equation}

As BIC is expressed in terms of MSE as described above and as DRBM makes the cluster separation linearly separable as observed in Fig. \ref{Wine}(b), it becomes easy for BIC to predict the number of clusters. SOM uses the Gaussian neighborhood function to find the nearest neighbors and uses Euclidean distance to find the winning neuron. Hence, SOM works well for datasets which are more linearly separable and for datasets where the clusters are non-overlapping. Hence, applying a DRBM transformation on the existing dataset facilitates SOM to calculate the cluster centroids more accurately which in turn increases SOM clustering efficiency.

\section{Acknowledgement}
This research is supported by the Accelerated Materials Development for Manufacturing Program at A*STAR via the AME Programmatic Fund by the Agency for Science, Technology and Research under Grant No. A1898b0043.

\bibliographystyle{IEEEtran}

\end{document}